\documentclass[twoside]{article}

\usepackage{amssymb}
\usepackage{subfigure}
\usepackage{graphicx}
\usepackage{makecell}
\usepackage{multirow}

\usepackage[accepted]{aistats2021}
\usepackage{url}
\usepackage{hyperref}
\hypersetup{
     colorlinks = true,
     linkcolor = blue,
     anchorcolor = blue,
     citecolor = blue,
     filecolor = blue,
     urlcolor = blue
     }

\usepackage[round]{natbib} 

\usepackage{booktabs}
\usepackage{multirow}
\usepackage[table,xcdraw]{xcolor}

\begin{document}

%
\runningtitle{Learning \textbf{\textit{Matching}} Representations for Individualized Organ Transplantation Allocation}

%
\runningauthor{C. Xu, A. M. Alaa, I. Bica, B. D. Ershoff, M. Cannesson, M. van der Schaar}

\twocolumn[

\aistatstitle{Learning \textbf{\textit{Matching}} Representations for Individualized \\ Organ Transplantation Allocation}

\aistatsauthor{Can Xu$^*$ \\ University of Cambridge \And Ahmed M. Alaa$^*$ \\ UCLA \AND Ioana Bica \\ University of Oxford \\ The Alan Turing Institute \And Brent D. Ershoff \\ UCLA \And Maxime Cannesson \\ UCLA \And Mihaela van der Schaar \\ University of Cambridge \\ UCLA \\ The Alan Turing Institute}

\aistatsaddress{}
]

\begin{abstract}
Organ transplantation is often~the~last~resort for treating end-stage illness, but the~probability of a successful transplantation depends greatly on {\it compatibility} between donors and recipients. Current medical practice relies on coarse rules for donor-recipient matching, but is short of domain knowledge regarding the complex factors underlying organ compatibility. In this paper, we~formulate~the problem of learning {\it data-driven}~rules~for organ matching using observational data for organ allocations and transplant outcomes.~This~problem departs from the standard supervised~learning setup in that it involves matching~the~two feature spaces (i.e., donors and~recipients), and requires estimating transplant outcomes under counterfactual matches {\it not} observed in the data. To address~these~problems,~we propose a model based on {\it representation learning} to predict donor-recipient compatibility; our model learns representations that cluster donor features, and applies donor-invariant transformations to recipient features to predict outcomes for a given donor-recipient feature instance. Experiments on semi-synthetic and real-world datasets show that our model outperforms state-of-art allocation~methods and policies executed by human experts.
\end{abstract}

\section{INTRODUCTION}
\label{Sec1}
Organ transplantation is the definitive therapy~for~patients with end-stage diseases who are~unresponsive~to medical therapies \citep{lechler2005organ}.~Whereas~organ transplantation can improve~life~expectancy~and~quality of life for recipients, the risks of transplant failure and/or post-operative complications (including infections, chronic rejection and malignancy) are also significant \citep{rubin2002infection}. These risks depend greatly on the {\it compatibility} between the clinical characteristics of recipients and donors, hence pre-operative anticipation of organ compatibility is key for proper donor-recipient matching and organ allocation \citep{briceno2013donor}.

\begin{figure}
    \centering
    \includegraphics[width=2.5in]{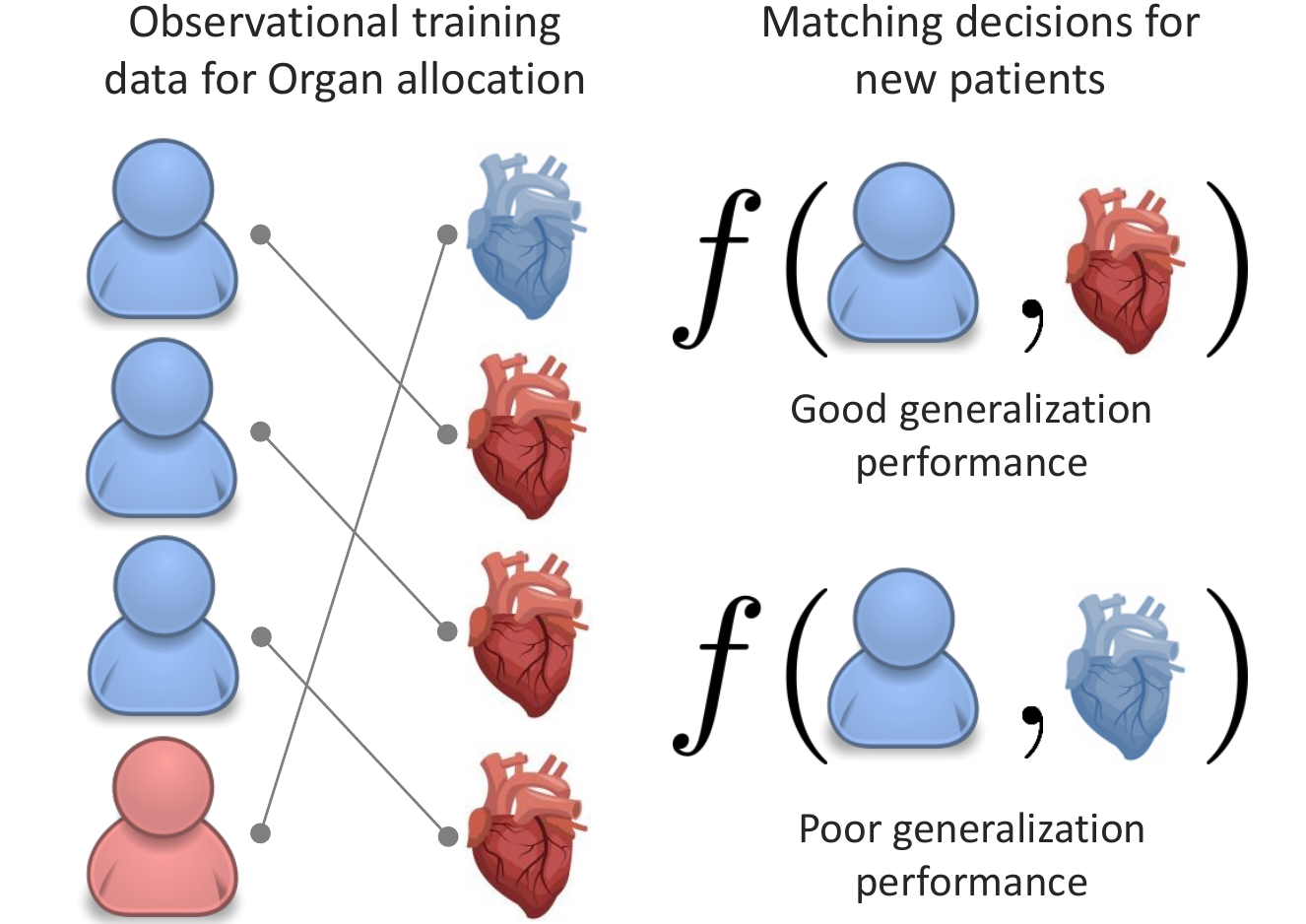}
    \caption{\footnotesize {\bf Donor-recipient matching for organ~transplantation.} We~show~an~exemplary~training data set with two types of donors and recipients (red and blue in either case). In current practice, blue recipients are consistently allocated red organs, and vice versa; hence we only observe blue-red and red-blue matches in training data. Using a supervised learning model $f$ to predict transplant outcomes for alternative allocations would provide accurate predictions for blue-red/red-blue matches but would generalize poorly to blue-blue/red-red matches, making it challenging to learn new allocation rules from the training data.}
    \rule{\linewidth}{.75pt}
    \label{Fig3}
    \vspace{-.4in}
\end{figure}

While some of the clinical factors pertaining to organ compatibility are already known, e.g., blood and tissue types \citep{delmonico2004donor}, it is hypothesized that donor-recipient compatibility involves additional clinical factors and exhibits a much more intricate pattern of feature interaction \citep{gentry2011kidney}. Uncovering such factors and patterns using~organ~allocation trials is infeasible and~unethical--~this~motivates a data-driven approach for learning donor-recipient compatibility using (historical) observational~data~for~organ~allocations and transplant outcomes.  

In this paper, we formulate the learning problem of estimating organ compatibility using observational data for previous donor-recipient matches.~This~problem~departs from the standard supervised learning setup in two ways. First, it involves learning a function that is defined over \textit{two} feature spaces (for donors and recipients), one of which (the donor feature) can be thought of as an {\it interventional} variable. Second, learning the compatibility function requires estimating the transplant outcomes under {\it counterfactual} matches not observed in the data (Figure \ref{Fig3}). This problem also departs from the treatment effect estimation setup (e.g., \citep{alaa_bayesian_2017, alaa_limits_2018, shalit_estimating_2017, yao_representation_2018, zhang_learning_2020}) in that the interventional variable (donor features) is potentially continuous-valued and high-dimensional, which render existing solutions based on the binary potential outcomes framework inapplicable. 

To address this problem, we propose a~model~based~on {\it representation learning} to predict donor-recipient compatibility. The proposed representation addresses~the problem of the high-dimensional donor feature space by clustering all donors into a set of~donor~``types''---it then applies a {\it donor-invariant}~transformation~to~the recipient features to predict outcomes for a given~instance of a donor-recipient match.~The~donor-invariant representation is learned by minimizing the probability distance between the feature distribution of all recipients and those of recipients matched with each type of donor. This ``domain-adaptation'' approach enables the overall predictive model to generalize well when tested on multiple potential matches for new patients, including matches rarely observed in the data. We call this approach: {\it matching representation} learning.

We use a deep embedded clustering network \citep{xie2016unsupervised} to learn the donor type clusters and a standard feed-forward network to learn the donor-invariant representations. The features learned by our matching representation are then passed to a multi-headed predictive neural network that predicts the transplant outcome for a given recipient under all possible donor types---the entire model is jointly trained end-to-end. As we show in Section \ref{Sec4}, experiments datasets demonstrate that our model outperforms state-of-art allocation methods and policies executed by human experts.

{\bf Ethical considerations.} The problem of allocation and prioritization of scarce donor organs to terminally-ill patients is associated with ethical issues \citep{abouna2003ethical}. The goal of this work is to provide tools for {\it understanding} organ compatibility and {\it informing} matching decisions rather than fully {\it automating} the process.

\section{PROBLEM FORMULATION}
\label{Sec2} 
\subsection{Organ transplantation data}
\label{Sec21}
Let $\mathcal{X}_r$ and $\mathcal{X}_o$ be two feature spaces~of~dimensions~$d_r$ and $d_o$, respectively. Let $x_r \in \mathcal{X}_r$~and~$x_o \in \mathcal{X}_o$~be~two feature instances for a recipient~and~an~organ~donor~---~our key objective is to estimate a {\it compatibility function} $C(x_r, x_o): \mathcal{X}_r \times \mathcal{X}_o \to \mathcal{Y}$ that maps the recipient and donor features to a transplant outcome $y \in \mathcal{Y}$. The transplant outcome $y$ can be defined as the transplant success probability or post-transplant survival time. 

Typically, we are presented with an observational data set $\mathcal{D}_n$ comprising $n$ pairs of recipients and donors, i.e.,
\begin{align}
\mathcal{D}_n \triangleq \left\{x^{(i)}_r, x^{(i)}_o, y^{(i)}\right\}^n_{i=1},
\label{Sec2eq1}
\end{align}
and our learning task entails estimating $C(x_r, x_o)$ using the samples in $\mathcal{D}_n$. Note that each donor-recipient pair $(x^{(i)}_r, x^{(i)}_o)$ in $\mathcal{D}_n$ is matched according to an underlying process that depends on donor organ availability and existing clinical guidelines on organ matching criteria, e.g., blood and tissue types \citep{israni2014new}. However, donors and recipients in $\mathcal{D}_n$ could have been matched {\it differently}---~an~accurate~estimate of the compatibility function, $C(x^{(i)}_r, x^{(j)}_o), \forall i \neq j,$ may inform clinicians of an alternative match that would have improved patient outcomes.~(For~notational~brevity,~we drop the superscript $i$ in the remainder of the paper.)   

\subsection{Learning to \textbf{\textit{match}} with little supervision}
\label{Sec22} 
The (joint) distribution of donors~and~recipients~in~$\mathcal{D}_n$ depends on: the distribution of patients~$\mathbb{P}(x_r)$,~in~addition to the distribution of available~donors~along~with the underlying matching policy, both absorbed into a conditional {\it matching} distribution $\mathbb{P}(x_o\,|\,x_r)$, i.e., 
\begin{align}
\mathbb{P}(x_o, x_r) = \mathbb{P}(X_o=x_o\,|\,X_r=x_r) \cdot \mathbb{P}(X_r=x_r). \nonumber
\end{align}
Note that the matching distribution $\mathbb{P}(X_o=x_o\,|\,X_r=x_r)$ is not known ahead of time---it depends on the organs available to recipients at the time of their surgery and the process (clinical guidelines) by which clinicians match donor organs and recipients. 

Now consider an estimate $\widehat{C}$ of the compatibility function $C$---the loss function associated with $\widehat{C}$ is: 
\begin{align}
L(\widehat{C}) = \mathbb{E}_{X_r} \mathbb{E}_{X_o}\ell(y, \widehat{C}(X_r, X_o)),
\label{Sec2eq2}
\end{align}
where $\ell(.)$ is the loss associated with a donor-recipient feature instance, and $L$ is the expected loss for $\widehat{C}$.  

We note that the loss function in (\ref{Sec2eq2}) is {\it independent} of the matching distribution $\mathbb{P}(x_o\,|\,x_r)$.~That~is,~we~average the instance-wise loss $\ell(y, \widehat{C})$ over~the~{\it marginal}~distributions of donors and recipients, $\mathbb{P}(x_o)$ and $\mathbb{P}(x_r)$, rather than their joint~distribution~$\mathbb{P}(x_o, x_r)$.~This is because our estimate $\widehat{C}$ is meant to be used to predict transplant outcomes under {\it alternative} matches different from the ones observed in the data. In other words, we want our estimate $\widehat{C}$ to generalize well for all possible pairs of donors and recipients, not just the pairs that are frequently matched in the observational data.  

\begin{figure}
    \centering
    \includegraphics[width=2in]{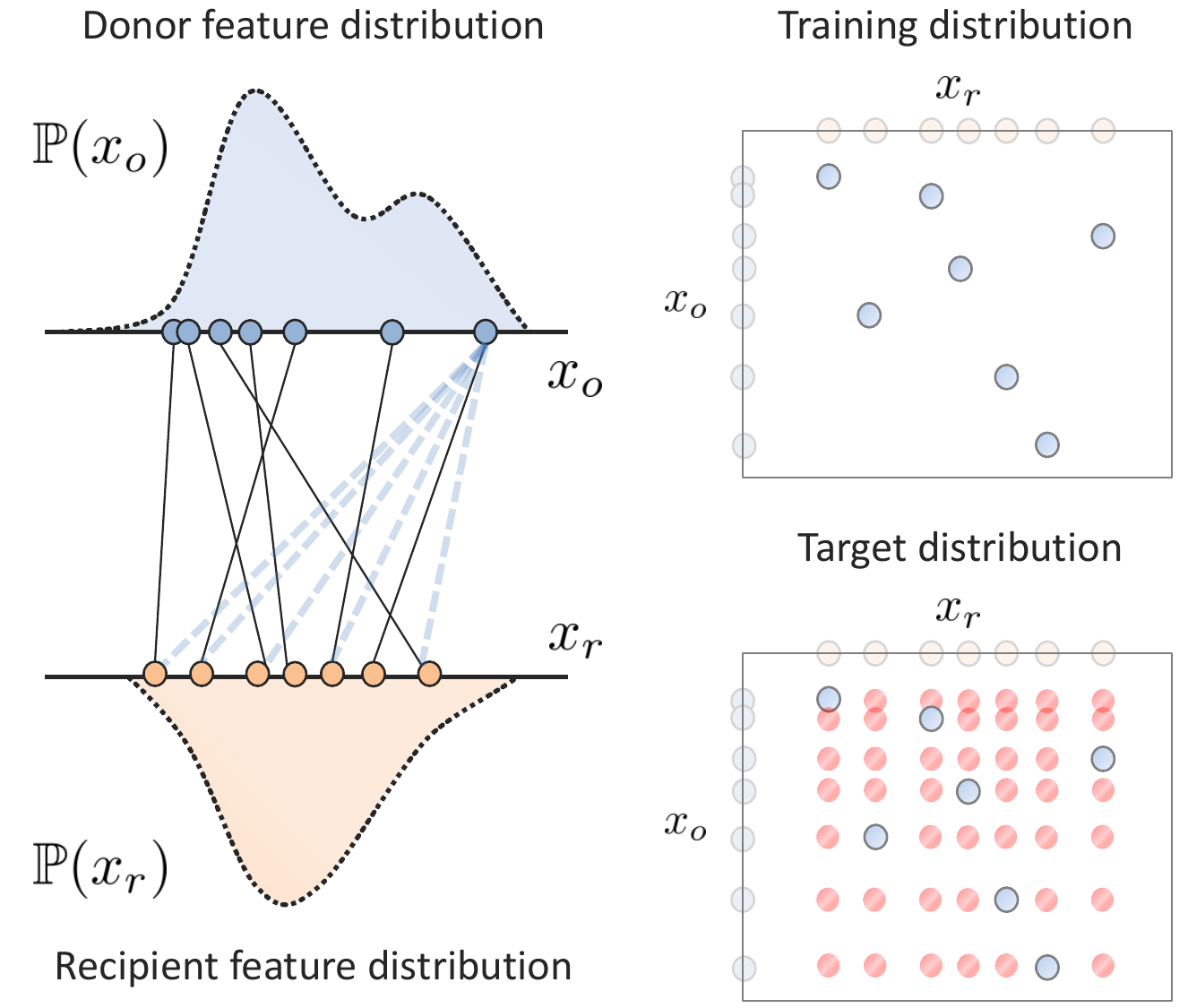}
    \caption{\footnotesize {\bf Illustration for the donor-recipient matching problem.} Here, solid black lines correspond~to~factual matches and dotted blue ones correspond to counterfactual matches (left). The right panel shows the empirical training/target distributions---blue dots correspond to observed transplant outcomes whereas red dots are transplant outcomes for all of the (unobserved) counterfactual matches.}
    \rule{\linewidth}{.75pt}
    \label{Fig1}
    \vspace{-.25in}
\end{figure}

Why~is~learning~the~compatibility~function~$C(x_r, x_o)$ {\it not} a~standard~supervised~learning~problem?~The~reason we cannot simply use supervised methods to learn $C$ is because of the discrepancy between the distribution of donor-recipient pairs in the {\it training}~data,~and the {\it target} distribution of donor organs and recipients to which the model will be practically applied, i.e., 
\begin{align}
\textbf{Training distribution:}&\,\,\, \mathbb{P}(X_r, X_o), \nonumber \\
\textbf{Target distribution:}&\,\,\, \mathbb{P}(X_r) \cdot \mathbb{P}(X_o). \nonumber
\end{align}
That is, we want to train the~model~on~data~samples~of (already-matched) donor-recipient pairs, but then apply the model to predict outcomes of all possible donor-recipient matches in order to inform matching decisions (See Figure \ref{Fig1}). If we are to use supervised learning via empirical loss minimization, the empirical risk $\widehat{L}$ for an estimate $\widehat{C}$ would be
\begin{align}
\widehat{L}(\widehat{C}) = \sum_{i}\ell(y^{(i)}, \widehat{C}^{(i)}) + \sum_{i\neq j}\underbrace{\ell(y^{(i,j)}, \widehat{C}(x^{(i)}_r, x^{(j)}_o))}_{\mbox{\footnotesize Counterfactual match}},
\nonumber
\end{align}
where $y^{(i,j)}$ is the true transplant outcome for recipient $i$ having been given donor $j$'s organ. Thus,~evaluating the empirical loss $\widehat{L}(\widehat{C})$ entails evaluating the instance-wise loss of all the $n^2$ possible matches between donors and recipients. However,~we~only~observe~$n$~``factual'' matches in $\mathcal{D}_n$; the remaining $n^2-n$~donor-recipient matches are ``counterfactual'', and so we do not observe any transplant outcome $y^{(i,j)}$ except~for~$i=j$, rendering empirical risk minimization infeasible.  

\begin{figure*}[t]
  \centering
  \includegraphics[width=5.5in]{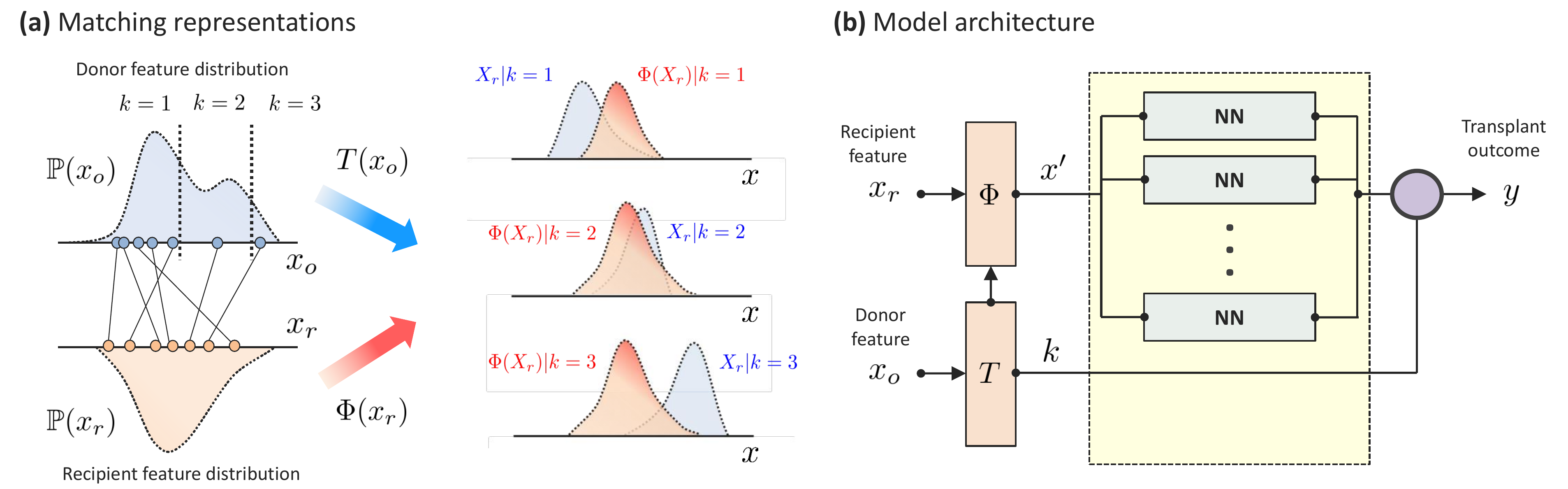}
  \caption{\footnotesize {\bf Pictorial depiction of matching representation.} (a) Here, we show the same~distributions~for~donor~and~recipient features as in Figure \ref{Fig1}. (We show one-dimensional feature spaces $\mathcal{X}_r$ and $\mathcal{X}_o$ for simplicity of exposition.) The donor type map $T(.)$ clusters the donor feature space into 3 distinct clusters. Because the matching distribution~$\mathbb{P}(X_o|X_r)$~is~not equal to $\mathbb{P}(X_o)$, the conditional distribution of recipient features given the matched donor types~are~different.~(For~example, the age distribution for patients receiving kidneys from donors with blood type A is different from that of~patients~matched to donors with blood type B.) The mapping $\Phi$ transforms $X_r$ to a random variable $X^\prime$ with a distribution that is invariant to the donor type. (b) Architecture for our model. The recipient feature is passed through the mapping $\Phi$~and~the~donor feature is passed through $T$, both implemented through a feed-forward neural network. The output of $\Phi$, $x^\prime$, is passed through a multi-headed neural network, with each head predicting the transplant outcome under different donor types.} 
  \rule{\linewidth}{.75pt}    
  \label{Fig2} 
  \vspace{-6mm} 
\end{figure*} 

\section{MATCHING REPRESENTATIONS}
\label{Sec3}
As we have seen in Section \ref{Sec2}, learning the compatibility function $C(x_r, x_o)$ via empirical risk minimization is not possible as the empirical estimate of (\ref{Sec2eq2})~is inaccessible. To address this challenge,~we~develop~a~model for estimating $C$ using {\it matching representations} that neutralize the bias induced by the matching distribution $\mathbb{P}(x_o\,|\,x_r)$, and generalizes well to the~marginal~distributions of all donors and recipients.   

\subsection{Match-Invariant Representations}
\label{Sec31}
The discrepancy between the~training~(or~source)~and target distributions renders our learning~problem~akin to {\it domain adaptation}; a~key~technique~in this setup is the usage of {\it domain-invariant} representations to alleviate the generalization errors resulting from distribution mismatches \citep{shalit_estimating_2017, zhao2019learning}.

Building on the concept of domain-invariance, we define a {\it match-invariant} representation ${\bf\Phi}: \mathcal{X}_r \times \mathcal{X}_o \to \mathcal{Z}$ as one that satisfies the following condition:  
\begin{align}
\mathbb{E}_{\mathbb{P}(X_r, X_o)}\, f({\bf\Phi}(X_r, X_o)) = \mathbb{E}_{\mathbb{P}(X_r)\, \mathbb{P}(X_o)} f({\bf\Phi}(X_r, X_o)),
\nonumber
\end{align}
for some function $f$.~That~is,~the~representation~${\bf\Phi}$~alleviates the (confounding) effect of the matching distribution $\mathbb{P}(x_o\,|\,x_r)$, enabling models trained on transformed donor-recipient features, ${\bf\Phi}(X_r, X_o)$, to generalize to the target distribution. The match-invariance condition plays a key role in the matching representations that we construct throughout this Section.  

\subsection{Latent Donor Types}
\label{Sec32}
Constructing a match-invariant representation ${\bf\Phi}$~is~a complicated task, especially when the donor and recipient feature spaces, $\mathcal{X}_r$ and $\mathcal{X}_o$, are high-dimensional. To simplify the design of ${\bf\Phi}$, we assume that donors belong to a set of $K$ ``types''.~Let~$T: \mathcal{X}_o \to \{1,\ldots, K\}$~be a map from donor features to discrete donor types; the matching distribution with respect to donor types can be given by $\mathbb{P}(T(x_o)=k\,\,|\,x_r),\, k \in \{1,\ldots, K\}$. 

The donor type map $T$ is {\it not} predefined ahead of time, but is rather learned from the data as part of the representation ${\bf\Phi}$. The clustering of donors is not only useful for simplifying the construction of matching representations, but also for providing~interpretable~grouping of donor types that could be conveniently incorporated in clinical guidelines \citep{barr2006report}.  

\subsection{Building Matching Representations}
\label{Sec33}
We propose the following construction~for~a~matching representation ${\bf\Phi}: \mathcal{X}_r \times \mathcal{X}_o \to \mathcal{Z}$ that~satisfies~the invariance condition in Section \ref{Sec31}. The proposed representation converts the feature pair $(x_r, x_o) \in \mathcal{X}_r \times \mathcal{X}_o$ to $z=(x^\prime, k) \in \mathcal{Z}$ through the transformation  
\begin{align}
{\bf\Phi}(x_r, x_o) = (x^\prime, k) = (\Phi(x_r),\, T(x_o)).
\label{Sec33eq0} 
\end{align} 
The mapping in (\ref{Sec33eq0}) jointly transforms the donor feature $x_o$ to donor type $k$, and recipient~feature~$x_r \in \mathcal{X}_r$ to $x^\prime \in \mathcal{X}^\prime$. The mapping $\Phi: \mathcal{X}_r \to \mathcal{X}^\prime$~is~designed to ensure that ${\bf\Phi}$ satisfies match-invariance, which in the case of a clustered donor feature space reduces to:
\begin{align}
\mathbb{E}_{X_r}\, f(\Phi(X_r)) = \mathbb{E}_{X_r|T(X_o)=k}\, f(\Phi(X_r)),
\label{Sec33eq1} 
\end{align} 
for all $X_o \in \mathcal{X}_o,\, T(X_o)=k,\, \forall k \in \{1,\ldots, K\}$.~That~is, the mapping $\Phi$ transforms $X_r$ to a new feature space wherein the distributions of $\Phi(X_r)\,|\,T(X_o)=k$ are the same for all latent donor types $k \in \{1,\ldots, K\}$. 

Figure \ref{Fig2}(a) pictorially visualizes the matching representation ${\bf\Phi}$ in the donor-recipient~feature~space.~As~we can see, ${\bf\Phi}$ discretizes the donor feature space $\mathcal{X}_o$ into three distinct clusters (donor types). The distribution of recipients $X_r$ matched to organs~from~each~donor cluster differ significantly---the mapping~$X^\prime = \Phi(X_r)$ transforms $X_r$ to a new feature space where the distributions $X^\prime\,|\,T(X_o)=k, \forall k \in \{1, 2, 3\}$ are similar.

\subsection{Learning \textit{compatibility} functions with matching representations}
\label{Sec34}
Having constructed a matching representation for the donor-recipient features, we estimate the compatibility function $C(x_r, x_o)$ in the transformed feature space $\mathcal{Z}$ rather than $\mathcal{X}_r \times \mathcal{X}_o$ as follows:
\begin{align}
\widehat{C}(x_r, x_o) = f({\bf\Phi}(x_r, x_o)) = f(x^\prime, k),
\label{Sec34eq1} 
\end{align} 
where $f:\mathcal{Z}\to\mathcal{Y}$ is a predictive~model~that~maps~the transformed (donor-recipient) features $z$ to transplant outcome $y$. Thus, our model comprises two learnable components: the predictive model $f$, and~the~matching representation ${\bf\Phi}$. In what follows,~we provide the model specification for $f$ and ${\bf\Phi}$. 

{\bf Multi-headed predictive~network.}~Given~a~matching representation, we model~$f({\bf\Phi}(x_r, x_o))=f(x^\prime, k)$ as a multi-headed neural network model, where we~use $K$ feed-forward neural networks to predict the transplant outcome under each type of donor~for~a~given~recipient with a feature $x_r$ (See Figure \ref{Fig2}(b)). We denote the predictive loss associated with $f$ as $L_f$. 

{\bf Donor type mapping via deep embedded clustering.} Clustering the donor space is an unsupervised problem as the donor types are not defined or labeled. We use a Deep Embedded Clustering~(DEC) network \citep{xie2016unsupervised} to model the donor type mapping $T(x_o)$. Our choice of DEC is motivated by~the~ease~of its incorporation into an end-to-end training procedure that involves the other model components ($\Phi$ and $f$). 

DEC involves training an autoencoder at first to learn a representation of donor features.~The~encoder~part~of the autoencoder is then used to learn a clustering using the following loss function:
\begin{align}
    t_{ij} &= \frac{(1 + ||{\bf d}_i - \mathbf{\mu}_j||^2)^{-\frac{1}{2}}}{\sum_j (1 + ||\mathbf{d}_i - \mathbf{\mu}_j||^2)^{-\frac{1}{2}}},\,\,
    p_{ij} = \frac{\frac{t_{ij}^2}{\sum_i t_{ij}}}{\sum_j \frac{t_{ij}^2}{\sum_i t_{ij}}}
    \notag\\
    L_{DEC} &= \sum_i \sum_j p_{ij} \log \frac{p_{ij}}{t_{ij}}
\end{align}
where $\mathbf{d}_i$ is the learned representation of donor $i$ (produced by the encoder), $\mathbf{\mu}_j$ is a cluster center (which is randomly initialized), and $t_{ij}$ represent the probability of donor $i$ belonging to cluster $j$.

{\bf Matching representations using probability distance minimization.} Finally, the last component of our model is the feature map $\Phi:\mathcal{X}_r \to \mathcal{X}^\prime$, which we model via a (feed-forward) neural network. To achieve the match-invariance condition in (\ref{Sec33eq1}), we learn $\Phi$ by minimizing the distance among distributions of representations of recipients assigned to different types of donors through the following loss function: 
\begin{align}
L_{\Phi} = \sum^K_{k=1}d(\mathbb{P}_{X_r}(\Phi), \mathbb{P}_{X_r|T(X_o)=k}(\Phi)), 
\label{Sec34eq2} 
\end{align} 
where $\mathbb{P}(\Phi) = \mathbb{P}(\Phi(X_r))$ ($X_r$ is dropped for notational brevity), and $d(., .)$ is~a~probability~distance~metric, such as the integral probability~metrics~(IPM) \citep{mullerintegral1997}, total variation distance or KL divergence. Here, we use the KL divergence metric~assuming~that~the~distributions of $X_r$ and $X_r\,|\,T(X_o)$ are Gaussian.  

The overall model is trained~end-to-end~by~combining the loss functions $L_f$, $L_{DEC}$, and $L_{\Phi}$, as follows:
\begin{align}
    L = \sum L_f + \alpha \cdot L_{DEC} + \beta \cdot L_{\Phi}
\end{align}
where $\alpha$ and $\beta$ are hyperparameters balancing losses of different components of the model.~We~also~treat~the number of donor types $K$ as a hyperparameter.

\section{RELATED WORKS}
\label{Sec4}
Our work relates to two strands of previous literature: (1) machine learning-based models for transplantation, and (2) models for estimating treatment~effects.~In~this Section, we explain how our work relates~to~these.

{\bf Machine learning-based models for organ transplantation.} Previous works on organ transplants focus on developing a more accurate risk model for predicting survival after transplant \citep{medved_improving_2018, nilsson_international_2015}. In \citep{nilsson_international_2015}, a deep neural network is proposed with classification and regression trees to predict transplantation outcomes and evaluate the impact of recipient-donor variables on survival. Instead of improving the accuracy of prediction of survival, other works focus on improving recipient-donor matching. For instance,~\citep{yoon_personalized_2017} partition recipient-donor feature space into subspaces and use a separate prediction model for each subspace. In this architecture, each independent prediction model is trained to solve a more specific and less general sub-problem. Therefore, models for subspaces of matched recipient-donor pairs are expected to be more robust than models trained to~solve~the~general problem. Unlike our model, this approach does not handle the matching bias in the data, hence it cannot be reliably used to recommend alternative matches other than those observed in the data.

\noindent
{\bf Estimating treatment effects.}~The~problem~of~estimating the effects of treatments from observational data shares similarities with our setup---in this setup, the effects of a treatment are estimated~by~inferring~its counterfactual outcomes while accounting for the~{\it selection bias} resulting from the data being generated~according to an underlying treatment policy. A typical approach in existing literature is to fit a single model to estimate all counterfactuals outcomes of a treatment, and distributions of different treatments' populations are adjusted (balanced) to handle selection bias. For instance, \citep{wager_estimation_2018} uses random forest, and \citep{johansson_learning_2016, shalit_estimating_2017} use deep neural networks to solve treatment effects estimation problem under this single model methodology. On the contrary, \citep{alaa_bayesian_2017, alaa_deep_2017} use multi-task approaches that are analogous to our multi-headed predictive network, such as multi-task Gaussian process, to estimate treatment effects. Our proposed model is most similar to \citep{alaa_deep_2017} and \citep{shalit_estimating_2017}, since in all these works, deep neural networks with multiple heads are used to estimate potential outcomes. However, these works focus on binary treatment effects, hence they cannot handle the continuous, high-dimensional interventions in our setting (donor features). 
 
\begin{figure*}[h]
    \centering
    \subfigure[{\footnotesize Distribution of recipients.}]{\includegraphics[width=.28\linewidth]{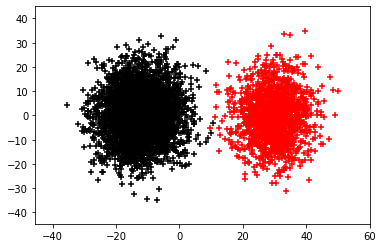}}
    \subfigure[{\footnotesize Distribution of donors.}]{\includegraphics[width=.28\linewidth]{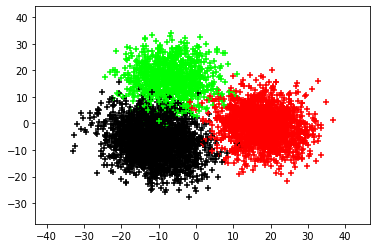}}
    \subfigure[{\footnotesize Donor-recipient pairs.}]{\includegraphics[width=.28\linewidth]{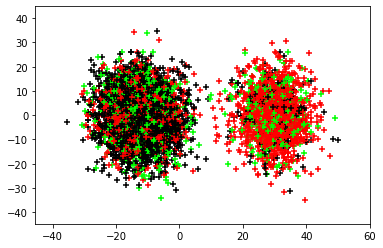}}
    \caption{{\footnotesize Distributions of recipient and donor features for the synthetic model in Section \ref{Sec51}.~In~(a),~\textbf{Black}~denotes~$m=1$ and \textbf{\textcolor{red}{Red}} denotes $m=2$. In (b) and (c), \textbf{Black} denotes $k=1$, \textbf{\textcolor{red}{Red}} denotes $k=2$, and \textbf{\textcolor{green}{Green}} denotes $k=3$.}}
  \rule{\linewidth}{.75pt}    
  \label{Fig4} 
  \vspace{-4mm} 
\end{figure*}


\section{EXPERIMENTS}
\label{Sec5}
Since we never observe all counterfactual matches in the real world, there is no ground truth target (label) to evaluate estimated donor-recipient compatibility in matches other than the ones observed in~data.~Hence, it is difficult to evaluate the performances of models using only real-world data. Previous works on related problems involving counterfactual inference use semi-synthetic data comprising real covariates and simulated outcomes to evaluate performance (e.g., \citep{pearl_causality_2009, louizos_causal_2017, shalit_estimating_2017, yoon_ganite_2018}). In this Section, we use synthetic and semi-synthetic data sets to quantitatively evaluate the performance of our proposed model. We also use real-world data sets to qualitatively test the performance of our model under real circumstances.

\subsection{Experiments on synthetic data}
\label{Sec51}
We evaluate the proposed matching representation on a synthetic data set to demonstrate its ability to discover the ``optimal'' clustering of donor types and better allocation policies from observational data.

We generate the synthetic data from a Gaussian mixture model, with a mixture of 2 recipient~types~and~3 donor types---we denote the recipient type as $m$ and the donor type as $k$. Recipients of different types have different probabilities of pairing with donors of different types as follows: 
\begin{table}[h]
    \centering
    \begin{tabular}{|c||c|c|c|}
    \hline
        $\mathbb{P}(k \mid m)$ & $k=1$ & $k=2$ & $k=3$ \\
        \hline\hline
        $m=1$ & 0.6 & 0.2 & 0.2 \\
        \hline
        $m=2$ & 0.1 & 0.7 & 0.2 \\
        \hline
    \end{tabular}
    \label{tab:my_label}
\end{table}
The data set contains selection bias since pairing of recipients from $m=2$ and donors from $k=1$ is rarely observed, hence the conditional distributions of recipients assigned to donor types are unbalanced. 

The transplantation outcome $y$ (survival time) associated with each donor-recipient match is modeled as:
\begin{align*}
    y \mid m=1 &\sim \mathcal{N}\left(\begin{bmatrix} 500 \\ 1000 \\ 1100 \end{bmatrix}, \begin{bmatrix} 50&0&0 \\ 0&100&0& \\ 0&0&100 \end{bmatrix}\right),
    \\
    y \mid m=2 &\sim  \mathcal{N}\left(\begin{bmatrix} 100 \\ 800 \\ 900 \end{bmatrix}, \begin{bmatrix} 10&0&0 \\ 0&100&0& \\ 0&0&100 \end{bmatrix}\right).
\end{align*}
The synthetic data is designed so that donor types 2 and 3 are ``similar'' and lead to similar transplantation outcomes for all recipients.~Besides,~recipients~of type 1 are more likely to be paired with donors of~type~1, despite the fact that donors of types 2 and 3 are better matches for the recipients. We sample 5,000 recipient-donor pairs and run several simulations on the synthetic data set to compare the allocation policy of our model and other state-of-art allocation policies. 

\begin{table*}[t]
    \centering
    \caption{{\footnotesize Simulation results of allocation policies evaluated on the synthetic data set. \textit{flipped ratio} stands for~the~ratio~of Type-1 recipients that are paired with Type-1 donors but now assigned to other type of donors. High \textit{flipped ratio} indicates that the policy can discover the better potential policy from the data set.}}
    {\footnotesize
    \vspace{.1in}
    \begin{tabular}{|l|ccc|cc|}
    \hline
        \multirow{2}{*}{\textbf{Policy}} & \multicolumn{3}{c|}{\textbf{All Recipients}} & \multicolumn{2}{c|}{\textbf{Type-1 Recipients}} \\
         & death rate & avg survival & avg benefit & flipped ratio & avg survival \\ 
        \hline\hline
        \textbf{FCFS} & 0.34 & 470.57 & 104.74 & n.a. & n.a. \\
        \textbf{UF} & 0.34 & 470.47 & 104.55 & n.a. & n.a. \\
        \textbf{BF} & 0.27 & 514.95 & 177.04 & n.a. & n.a. \\
        \hline
        \textbf{Real} & 0.34 & 474.28 & 104.10 & 0.00 & 508.97 \\
        \hline
        \textbf{Matching rep. (FCFS)} & 0.30 & 501.56 & 133.30 & 0.46 & 732.41 \\
        \textbf{Matching rep. (UF)} & 0.30 & 514.98 & 131.98 & 0.53 & 769.43\\
        \textbf{Matching rep. (BF)} & 0.28 & 528.43 & 165.57 & 0.41 & 710.50 \\
        \hline
    \end{tabular}}
    \label{tab:addexppolicy}
\end{table*}

We consider the following organ allocation policies:

(a) \textit{Real policy}: a new donor organ is allocated based on expert knowledge. This is the underlying policy by which the observational data was generated.

(b) \textit{First come first serve (FCFS)}: a new donor is always allocated to the first recipient in the waiting list.

(c) \textit{Utility first}: a new donor organ~is~allocated~to~the recipient with the best predicted transplantation outcome (survival time after transplantation).

(d) \textit{Benefit first}: a new donor organ is allocated~to~the recipient with highest benefit, where~benefit~is defined as the predicted gain in survival time after transplant. 

\begin{figure}[h]
    \centering
    \subfigure[Original clusters]{\includegraphics[width=.45\linewidth]{figures/biased_t_gt.png}}
    \subfigure[Trained clusters]{\includegraphics[width=.45\linewidth]{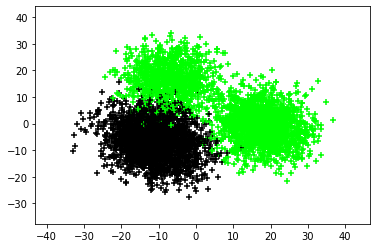}}
    \caption{{\footnotesize Visualization for the learned donor type clusters compared to the original clusters in the sampled data set. \textbf{Black} denotes $k=1$, \textbf{\textcolor{red}{Red}} denotes $k=2$, and \textbf{\textcolor{green}{Green}} denotes $k=3$. Donor types $Ok=2$ and $k=3$ are merged together in the jointly trained clustering model.}}
  \rule{\linewidth}{.75pt}    
  \label{Fig5} 
  \vspace{-5mm}
\end{figure}

We simulate these policies by examining the sampled donors and recipients in sequence and applying the allocation rules above. The performances of policies are evaluated in terms of recipients' average survival time after surgery, recipients' average net benefit in survival time of conducting surgery, and the death rate of recipients waiting in the sequence. A robust allocation policy is expected to have high average survival time and low death rate. Results are reported in Table~\ref{tab:addexppolicy}. We also visualize the clusters learned by our matching representations model in Figure \ref{Fig5}.

As shown in Figure \ref{Fig5}, the clustering~inferred~by~our matching representations merges the~2~``similar''~donor types into a single cluster, which improves the interpretability of the underlying data.~Additionally, as shown in Table \ref{tab:addexppolicy}, our model outperforms the baseline allocation policies in terms of the achieved survival time/rate. Matching representations achieve the highest average survival time as well as comparable death rate, compared with the best benchmark. In comparison with the real allocation policy of the dataset, it extended the average survival time by 11.4\%, the average benefit by 59\%, and reduced death rate by 6\%.

It is also noticable that about 50\% of Type-1 recipients originally assigned to Type-1 donors are assigned to donors of different types by our model. For this group of recipients, the flipping of assignments leads to a significant improvement of survival time for 51.2\%. The results suggest that our model is capable of discovering the better potential allocation policy from the observed data set, by clustering donors and constraining recipients to match the best type of donors. 

\begin{table*}[t]
\center
\caption{{\footnotesize Simulation results of allocation policies. Simulations are run on PLTSD and UNOS datasets.}}
\vspace{0.1in}
{\footnotesize
\begin{tabular}{@{}lccccccccc@{}}
\toprule
\multicolumn{1}{c}{\multirow{2}{*}{\textbf{Policy}}} & \multicolumn{3}{c}{\textbf{PLTSD}}                                                                                                                                         & \multicolumn{3}{c}{\textbf{UNOS-HR}}                                                                                                                                       & \multicolumn{3}{c}{\textbf{UNOS-LU}}                                                                                                                                       \\ \cmidrule(l){2-10} 
\multicolumn{1}{c}{}                                 & \begin{tabular}[c]{@{}c@{}}Death \\ rate\end{tabular} & \begin{tabular}[c]{@{}c@{}}Avg. \\ benefit\end{tabular} & \begin{tabular}[c]{@{}c@{}}Avg. \\ survival\end{tabular} & \begin{tabular}[c]{@{}c@{}}Death \\ rate\end{tabular} & \begin{tabular}[c]{@{}c@{}}Avg. \\ benefit\end{tabular} & \begin{tabular}[c]{@{}c@{}}Avg. \\ survival\end{tabular} & \begin{tabular}[c]{@{}c@{}}Death \\ rate\end{tabular} & \begin{tabular}[c]{@{}c@{}}Avg. \\ benefit\end{tabular} & \begin{tabular}[c]{@{}c@{}}Avg. \\ survival\end{tabular} \\ \cmidrule(r){1-10}
\textbf{FCFS}  &  0.25 & 700.75 & 1271.54 & 0.18 & 324.01 & 1959.47 & 0.18 & 213.52 & 1156.85\\
\textbf{UF}    &  0.24 & 
721.70 &
1295.21 &
0.21 &
203.35 &
2065.85 &
0.23 &
108.58 &
1210.71\\
\textbf{BF}  & 0.23 &
737.24 &
1306.01 &
0.16 &
377.71 &
2113.22 &
0.15 &
273.15 &
1288.77 \\
\textbf{Real}   &  0.25 &
703.23 &
1280.41 &
0.18 &
325.52 &
1974.74 &
0.18 &
219.46 &
1177.54  \\
\textbf{Matching rep. (FCFS)} &                                                      0.24 &
720.69 &
1285.09 &
0.16 &
357.20 &
2011.88 &
0.16 &
260.29 &
1193.32 \\
\textbf{Matching rep. (UF)} & 0.24 &
730.85 & 
1299.74 &
0.16 & 
320.53 & 
2057.54 & 
0.19 &
177.14 &
1216.79  \\
\textbf{Matching rep. (BF)}  &  0.23 &
736.98 & 
1305.16 &
0.16 &
389.69 &
2024.33 &
0.15 &
225.61 &
1224.48
 \\ \bottomrule
\end{tabular}}
\label{Tabnew3}
\end{table*} 
 
\vspace{-.05in} 
 
\subsection{Liver, lung and heart transplant data}
\label{Sec52}
Next, we test our model using three real-world datasets for organ transplantation. The first is the Paired Liver Transplant Standard Dataset {\bf (PLTSD)}, a database in which donor-recipient pairs are extracted from the \textbf{NHS Liver Transplant Standard Dataset}\footnote{\url{https://www.odt.nhs.uk/}}. This dataset contains 6,898 pairs of donors and recipients---recipients are characterized by 55 features, whereas donors have 28 features.~The~second~dataset~is~the~lung transplant data set from UNOS\footnote{\url{https://unos.org/data/}}, which comprises two sub-datasets: a set of 60,400 recipients who underwent heart transplantation, and 29,210 recipients who underwent lung transplantation over the years from 1987 to 2015, with 37 donor-recipient features in each. 

Because these are real-world data sets, we only observe transplant outcomes for factual~donor-recipient~matching but not for the counterfactual ones. Therefore, we simulate the transplant outcomes under counterfactual matches via a semi-synthetic model (See Appendix). We run experiments to evaluate the precision of estimated factual transplant outcomes, in addition to the precision of estimated potential outcomes under counterfactual matches. We use several evaluation metrics: (1) precision of estimated factual outcomes $\epsilon_F$; (2)  precision of estimated potential outcomes $\epsilon_{WMSE}$; (3) accuracy of the best donor type $AoDT$.
\begin{align}
    \epsilon_{F} &= \frac{1}{n} \sum_{i=1}^n ({\bf \hat{y}}^{(i)}[k^{(i)}] - y^{(i)})^2, \nonumber\\
    \epsilon_{WMSE} &= \frac{1}{n} \sum_{i=1}^n \boldsymbol{1}^\intercal_{1 \times K} \, ({\bf \hat{y}}^{(i)} - {\bf y}^{(i)})^2, \nonumber\\
    AoDT &= \frac{1}{n} \sum_{i=1}^n \max_k \hat{\mathbf{y}}^{(j)}[k],\nonumber
\end{align}
where ${\bf \hat{y}}^{(i)}[k^{(i)}] = \widehat{C}(x_r^{(i)}, T(x_o^{(i)})=k^{(i)})$ and ${\bf y}^{(i)}$ is a $1 \times K$ vector of outcomes under each donor type.

To our knowledge, there is no~existing~baselines~that has an equivalent problem formulation~to~ours.~Therefore, we use baselines sharing the same fundamental structure of our proposed model, where~there~is~a~clustering component to classify donors as well as a counterfactual estimation component to estimate potential outcomes. We use K-Means, EM, and DEC clusters as baselines for the clustering component. Baseline clusters are independent of the counterfactual estimation task, and they are not jointly trained with the counterfactual estimation component. For the predictive component, we utilize a linear regression and a multi-headed neural net as baselines. We also compare allocation policies (similar to those considered in Section \ref{Sec51}) based on our model to baselines based on several state-of-art regression models, namely Regression Neural Network (RegNN) \citep{specht_general_1991}, Regression Tree (RegTree) \citep{strobl_introduction_2009}, LASSO Regression \cite{tibshirani_regression_1996}, Ridge Regression \citep{hoerl_ridge_1970}, and ElasticNet Regression \citep{zou_regularization_2005}. Results are reported in Table \ref{tab:policy}. Data is divided into 90\%/10\% training/validating sets. Results are given in Tables \ref{Tabnew3}, \ref{tab:policy} and \ref{tab:main}.

\begin{table}
    \centering
    \caption{Simulation results of allocation policies. Simulations are run on PLTSD dataset.}
    \vspace{.1in}
    \tiny{
    \begin{tabular}{|l|ccc|}
    \hline
        \textbf{Policy} & \textbf{avg survival} & \textbf{avg benefit} & \textbf{death rate} \\
        \hline\hline
        Real & 1280.4136 & 703.234 & 0.2489 \\
        \hline
        FCFS & 1271.5453 & 700.754 & 0.2493 \\
        UF & 1295.2122 & 721.700 & 0.2419 \\
        BF & \textbf{1306.0093} & \textbf{737.245} & \textbf{0.2360} \\
        \hline
        RegNN & 1271.1415 & 700.516 & 0.2495 \\
        RegTree & 1276.1471 & 704.6127 & 0.2481 \\
        LASSO & 1293.5784 & 719.112 & 0.2445 \\
        Ridge & 1297.4642 & 723.233 & 0.2425 \\
        ElasticNet & 1283.6174 & 707.714 & 0.2470 \\
        \hline
        Matching rep. (FCFS) & 1285.0895 & 720.687 & 0.2431 \\
        Matching rep. (UF) & 1299.7383 & 730.849 & 0.2396 \\
        Matching rep. (BF) & \textbf{1305.1631} & \textbf{736.984} & \textbf{0.2366} \\
        \hline
    \end{tabular}}
    \label{tab:policy}
\end{table}

\begin{table}
    \center
    \caption{{\footnotesize Performance of all~baselines~on~PLTSD. The results reported are averaged over 50 trials.}}
    \vspace{.1in}
    \footnotesize{
    \begin{tabular}{|l|c|c|c|c|}
        \hline
        \textbf{Model} & $\epsilon_F$ & $\epsilon_{WMSE}$ & $AoDT$ \\
        \hline\hline
        K-Means/Linear & 15.5229 & 14.7163 & \\
        EM/Linear & 15.5218 & 14.7108 & \\
        DEC/Linear & 15.5227 & 14.7680 & \\
        \hline
        K-Means/NN & 14.6295 & 13.8614 & .5962 \\
        \hspace{1cm}with rep. & 14.6399 & 13.9016 & .5735\\
        EM/NN & 14.5779 & 13.9918 & .5504 \\
        \hspace{1cm}with rep. & 14.6302 & 13.9623 & .5567 \\
        DEC/NN & 14.5840 & 13.7966 & .5965 \\
        \hspace{1cm}with rep. & 14.5711 & 13.8430 & .6017 \\
        \hline
        Matching rep. & \textbf{14.5362} & 14.0140 & \textbf{.6260} \\
        \hline
    \end{tabular}}
    \label{tab:main}
\end{table}

As shown in Tables \ref{Tabnew3} and \ref{tab:policy}, our model shows comparable results to the best allocation policy BF, which significantly outperforms all other policies. Compared with the real policy, we observe that 5,224 out of the 6,898 (75.7\%) donors are assigned a different recipient by our model---in addition, our model extends the average survival time by 25 days (2\%) and the average benefit for 34 days (4.8\%). Our model also reduces the death rate by 4.9\%. 

As shown in Table \ref{tab:main}, our model outperforms benchmark models in many aspects. It is observable that our model produce better precisions with and without representation loss. For our proposed model, precision of estimated factual outcome improved by 3.4\% in PLTSD and FPLTSD respectively, compared to the best benchmark. As for $AoDT$, our model achieves the highest AoDT of 62.60\%, which improved by 4\% from the best benchmark and improved by 87.82\% compared to random guessing of donor-recipient matches.
 
\section{Conclusion}
In this paper, we developed a novel method for learning the compatibility between donors and recipients in the context of organ transplantation. The key challenge in this problem is that the underlying matching policies, driven by clinical guidelines, creates a ``matching bias'', and hence a co-variate shift in the joint distribution of donor and recipient features. To solve this problem, we developed a learning approach based on {\it matching representations} to learn a donor-recipient compatibility function that generalizes well to the marginal distributions of donors and recipients. Our approach learns feature representations by jointly clustering donor features, and applying donor-invariant transformations to recipient features to predict outcomes for a given donor-recipient instance. Experiments on multiple datasets show that our model outperforms state-of-art organ allocation methods.

\acknowledgments{
We would like to thank the reviewers for their invaluable feedback and the NHS Blood and Transplant for providing data. This work was supported by The Alan Turing Institute (ATI) under the EPSRC grant EP/N510129/1, The US Office of Naval Research (ONR), and the National Science Foundation (NSF).
}

\bibliographystyle{plainnat}
\bibliography{reference}

\end{document}